\documentclass[10pt, a4paper]{article}

\usepackage[final]{lrec-coling2024} 

\usepackage{multicol}
\usepackage{amsmath}
\usepackage{multirow}
\usepackage{booktabs}
\usepackage{makecell}
\usepackage{xcolor}
\usepackage{subfig}
\usepackage{hyperref}
 \definecolor{darkblue}{rgb}{0, 0, 0.5}
  \hypersetup{colorlinks=true, citecolor=darkblue, linkcolor=darkblue, urlcolor=darkblue}

\usepackage{xstring}
\usepackage[nameinlink,capitalize,noabbrev]{cleveref}

\newcommand{\norex}[1]{\textit{#1}}
\newcommand{\eng}[1]{`#1'}

\newtheorem{exmp}{Example}[section]

\usepackage{color}
\newtheorem{assumption}{Assumption}

\title{Estimating Lexical Complexity from Document-Level Distributions}

\name{Sondre Wold$^{*}$, Petter Mæhlum$^*$, Oddbjørn Hove} 

\address{$^*$University of Oslo, Helse Fonna \\
         sondrewo@ifi.uio.no, pettemae@ifi.uio.no, oddbjorn.hove@helse-fonna.no\\
         }

\abstract{
Existing methods for complexity estimation are typically developed for entire documents. This limitation in scope makes them inapplicable for shorter pieces of text, such as health assessment tools. These typically consist of lists of independent sentences, all of which are too short for existing methods to apply. The choice of wording in these assessment tools is crucial, as both the cognitive capacity and the linguistic competency of the intended patient groups could vary substantially. As a first step towards creating better tools for supporting health practitioners, we develop a two-step approach for estimating lexical complexity that does not rely on any pre-annotated data. We implement our approach for the Norwegian language and verify its effectiveness using statistical testing and a qualitative evaluation of samples from real assessment tools. We also investigate the relationship between our complexity measure and certain features typically associated with complexity in the literature, such as word length, frequency, and the number of syllables. 
 \\ \newline  }

\begin{document}

\maketitleabstract

\section{Introduction}
The assessment and diagnosis of mental health rely on interviews and questionnaires to ascertain the presence, severity, frequency, and duration of symptoms \citep{newson2020heterogeneity}. Assessments typically consist of lists of sentences of varying complexity and the successful conveyance of an item's intended meaning to a respondent can be influenced by both linguistic complexity and the types of judgments required to interpret the item. For instance, answering a question like, ``In the past 7 days, were you able to engage in conversations with various people, such as friends, students, and family?" demands that the respondent meet various linguistic requirements, including vocabulary knowledge and syntactic skills \citep{bell2018adapting, dolan2010universal}. To enhance the cognitive accessibility of assessments, it is recommended to follow specific guidelines that involve employing straightforward language and grammar in questions. However, it has been observed that this recommendation does not offer developers of assessment tools sufficiently concrete linguistic guidance \citep{kooijmans2022adaptation}. This motivates the development of better tools for estimating complexity to be used in health assessments. 

Estimates of text complexity have a long history within the educational sciences. Corpora such as the Common Core State Standards \citeplanguageresource{nelson2012measures} and the Standardized State Passage set (CCSS/ELA) \footnote{\url{https://learning.ccsso.org/common-core-state-standards-initiative}} have enabled research on how to use techniques from NLP in order to both quantify and predict the level of complexity for a given text. Many features of a text influence how accessible it is to a reader, such as vocabulary, style, and syntactic features \citep{hiebert2011beyond}. Previous work on text complexity has primarily focused on how to use such features to predict document-level complexity classes according to US age-grade levels. These levels can be used as supervision signals for typical learning algorithms \citep{flor-etal-2013-lexical, chen-meurers-2016-characterizing, sheehan-etal-2013-two}. 


In this article, we aim to develop estimates of complexity for Norwegian, a language for which there are no available 
corpora annotated for complexity. This limits the usage of available methods, such as supervised learning algorithms. Furthermore, we focus on the non-canonical task of estimating \textit{lexical} complexity, as more traditional document-level measures of complexity often require text lengths beyond what is typically used in assessment tools.

Our method is based on a two-step approach. We first collect four corpora of Norwegian texts that are assumed to belong to different levels of complexity: children's books, newspaper articles, encyclopedia entries written by domain experts for the general public, and legislative texts from the Norwegian parliament. We then verify that a document-level complexity metric called LIX produces complexity scores that separate documents into the four assumed complexity classes. In the second step, we calculate the median LIX score of the documents in which a word occurs, resulting in a complexity score for each lemma. We verify this procedure using both statistical testing and qualitative evaluation.

We believe that this procedure is an important first step towards developing better tools for improving the cognitive accessibility of mental health assessments. Our three main contributions are as follows: \emph{i)}~We develop a two-step approach for estimating lexical complexity that does not depend on any pre-annotated corpora, \emph{ii)}~we study how document-level distributions of complexity relate to word-level features, such as frequency, word length, and syllables, and \emph{iii)}~we release an interactive tool for suggesting alternative phrasings based on our findings.\footnote{\url{https://github.com/SondreWold/lexical_complexity_estimation}} 

\section{Methodology}
On a fundamental level, our approach rests on two key assumptions that are based on a version of the distributional hypothesis \citep{harris1954distributional}:

\begin{assumption}\label{assump:1}
Words of high \textit{lexical} complexity appear more frequently in \textit{documents} with high levels of complexity.
\end{assumption}

\begin{assumption}\label{assump:2}
If a document-level complexity measure is able to accurately separate documents that are \textit{known} to be of different complexity, then this metric contains information on the complexity of individual words within these documents. 
\end{assumption}

Given these assumptions, we formulate a two-step approach for quantifying the lexical complexity of a lemma given a text corpus. The following sections outline this procedure. We first show that a heuristic algorithm for document-level complexity, known as LIX, can segment texts from four different corpora into categories that match the assumed complexity of these documents. We then measure lexical complexity by looking at the median LIX score of the documents in which a lemma occurs.

\subsection{The LIX score}
\begin{table}[]
\normalsize
  \centering
    \begin{tabular}{@{}lc@{}}
    \toprule
    \textbf{Category} & \textbf{LIX}  \\
    \midrule
    Very easy & $20$ \\
    Easy & $30$ \\
    Medium difficulty & $40$ \\
    Difficult & $50$ \\
    Very difficult & $60$ \\
    \hline
    \end{tabular}
    \caption{Interpretation of LIX scores on a five-step scale of complexity.}
    \label{tab:lix}
\end{table}
Developed by Swedish scholar Carl-Hugo Björnsson, the LIX readability index weights the number of long words and the number of sentences against the total number of words in a text:

\begin{equation*}
    LIX = \frac{A}{B} + \frac{C*100}{A},
\end{equation*}
where $A$ is the number of tokens, $B$ the number of sentences and $C$ is the number of words with $>6$ letters. This word length was selected due to it giving the largest difference between simple and difficult texts \citep[p. 217]{bjornsson1968lasbarhet}. \citet[p. 89]{bjornsson1968lasbarhet} also lists an approximate distribution of scores according to reference points, which can be found in \cref{tab:lix}. It is important to note that this metric is designed and evaluated for Swedish. As Norwegian and Swedish are similar with respect to vocabulary construction, compounds, conjugation, and syntax, the LIX score is also applicable to Norwegian and there is a tradition of using it for assessing readability. 

It should be noted that we use the term \textit{complexity} where the LIX score originally refers to readability (swe.\textit{ l\"asbarhet}), and that the term can refer both to morphologically complex words or syntactically complex behavior or, as is the case in most work on Complex Word Identification (CWI), to the sum of all factors that make a word \textit{difficult} to understand for a certain group. \citet{bjornsson1968lasbarhet} himself uses the terms \textit{svårhetsgrad} `difficulty' and \textit{l\"attillg\"angelighet} `ease of access' as synonyms for readability. 

While LIX has been popular for Norwegian and Swedish, an example of a potential problem is the productivity of compounding as a morphological process. Compound words can be long ($>6$), but still be concrete and easily understandable, such as \textit{kosebamse} `teddy bear' (lit. \textit{hug bear}). Even at higher levels, long compound words can still be understandable. A morphologically complex word is not necessarily difficult, and a morphologically simple word might have a high lexical complexity. Examples from our data include \norex{art} \eng{species;type} with high complexity, and \norex{forstørrelsesglass} \eng{magnifying glass} with low complexity.

\subsection{Data collection}
\begin{table}[]
\normalsize
  \centering
    \begin{tabular}{@{}lcc@{}}
    \toprule
    \textbf{Dataset} & \textbf{\#}  & \textbf{LIX} \\
    \midrule
    Children & $3695$ &  $\mu=21.57, \sigma=4.56$ \\ 
    News & $111579$ &  $\mu=40.32, \sigma=5.82$ \\ 
    Encyclopedia & $17033$ & $\mu=45.40, \sigma=6.40$ \\ 
    Parliament & $2726$ & $\mu=47.04, \sigma=6.36$ \\ 
    \hline
    \noalign{\vskip 0.5mm}
    Total & $135033$ & $\mu=40.58, \sigma=6.94$ \\ 
   \bottomrule
    \end{tabular}
    \caption{Statistics for the different corpora with count (\#), mean ($\mu$) and standard deviation ($\sigma)$ after pre-processing and normalization.}
    \label{tab:corpora_statistics}
\end{table}


To evaluate that the LIX score can correctly distinguish between levels of complexity, we would ideally have liked to use something similar to the American Common Core State Standards \citeplanguageresource{nelson2012measures}. As there are no sufficiently large annotated resources on complexity in Norwegian that match our use case, we collect a wide range of texts and separate them into four discrete classes of complexity according to their source.  Some of this data is openly available, and some were provided to us under the condition that we only share aggregated statistics as the raw texts are protected by copyright. In addition to all code related to experiments, we release the aggregated statistics for reproducibility. A summary of the four corpora can be found in \cref{tab:corpora_statistics}.

\paragraph{Children's books}
We choose texts explicitly written for children to represent a low level of complexity. We collect statistics on 3\,695 books, both literary and non-fiction, written between 1950 and 2023, a period in which the written Norwegian language has been rather stable. Statistics about these texts are made available through an API from the National Library of Norway \footnote{\url{https://www.nb.no/dh-lab/}}. These texts are processed and digitized by the library using OCR. Although the raw texts are unavailable due to copyright protection, it is possible to extract the necessary information to calculate the LIX scores from this API, such as lemma lists and frequency tables. We provide scripts for executing the automatic extraction of this information. 

\paragraph{News articles}
We collect 111\,579 articles from the 2019 version of the Norwegian Newspaper Corpus. We include articles from ten different publications ranging from typical tabloids to more traditional prints, and specialized publications focusing on a single topic, like economics. The articles are publicly available as a single distribution. \footnote{\url{https://www.nb.no/sprakbanken/en/resource-catalogue/oai-nb-no-sbr-4/}}

\paragraph{Encyclopedia entries}
We collect 17\,033 texts from the Great Norwegian Encyclopedia (SNL).\footnote{\url{https://snl.no/}} SNL contains encyclopedia entries written by domain experts for the general public on a wide range of topics. The articles are collected by alphabetically traversing the lexicons sitemap and removing any markup. We were given explicit permission by editorial staff to crawl the website. As some of these texts are copied from printed books that have yet to fall under the public domain, we are limited to sharing aggregated statistics.

\paragraph{Texts from the Norwegian parliament}
As examples of documents with an assumed high level of complexity, we gather 2\,726 openly available legislative decision proposals from the Norwegian parliament. These are collected through their openly available API service. \footnote{\url{https://data.stortinget.no/}} We gather proposals from the year 2000 up until 2023.

\subsection{Preprocessing} \label{sec:pre_proc}

Texts from the news, encyclopedia and parliament corpora are tokenised, lemmatised and PoS-tagged using \textsc{stanza} \citep{qi2020stanza}. Due to copyright protections, we could not parse texts from the National Library ourselves but had to work on already processed texts. These were processed by research scientists at the National Library of Norway using \textsc{spaCy}. \footnote{\url{https://spacy.io/}} As the accuracy of both of these systems is comparable when evaluated on available benchmarks for Norwegian, we do not foresee any issues related to this. 

\paragraph{Written standards}
Norwegian has two written standards, Bokmål and Nynorsk. We do not distinguish between the two when collecting our corpora. Consequently, we assume that the complexity quantified by the LIX index produces similar scores for the two standards. However, it is not given that what one person would regard as a text of low complexity in Bokmål would be regarded as similar with respect to complexity if translated into Nynorsk. As the vocabulary, inflectional, and derivational morphology between the two are so similar with regard to the relevant features, we conjecture that it will not affect the LIX calculation, so the effect of separating the two is negligible. 

\subsection{Calculation of LIX scores}

We calculate the LIX score of the tokenized texts. Descriptive statistics about these scores can be found in \cref{tab:corpora_statistics}. For normalization of the data, we remove all outliers outside of four standard deviations. We also make sure that there are no samples with a LIX score above 100 (which is much higher than the typical range of this metric), as we found this to indicate that the parsing failed to correctly separate text elements that were delimited by the markup, e.g. tables.

\subsubsection{Distribution of scores}

As can be seen in figure \cref{tab:corpora_statistics}, the scores for the four corpora range from roughly 20 to 50, corresponding to a transition from ``very easy" to ``difficult", according to the taxonomy in \citet{bjornsson1968lasbarhet}. However, there is a considerable jump when moving from children's texts to news articles. The news, encyclopedia, and parliament corpora are all placed between the ``medium" to ``difficult" thresholds. We hypothesized that the texts from the parliament corpus would be closer to the threshold of being classified as ``very difficult" ($>=60$), although this turned out not to be the case. 

\subsection{Document-level complexity distributions} \label{sec:doc_level}
\begin{figure*}
    \centering
    \includegraphics[width=.7\textwidth]{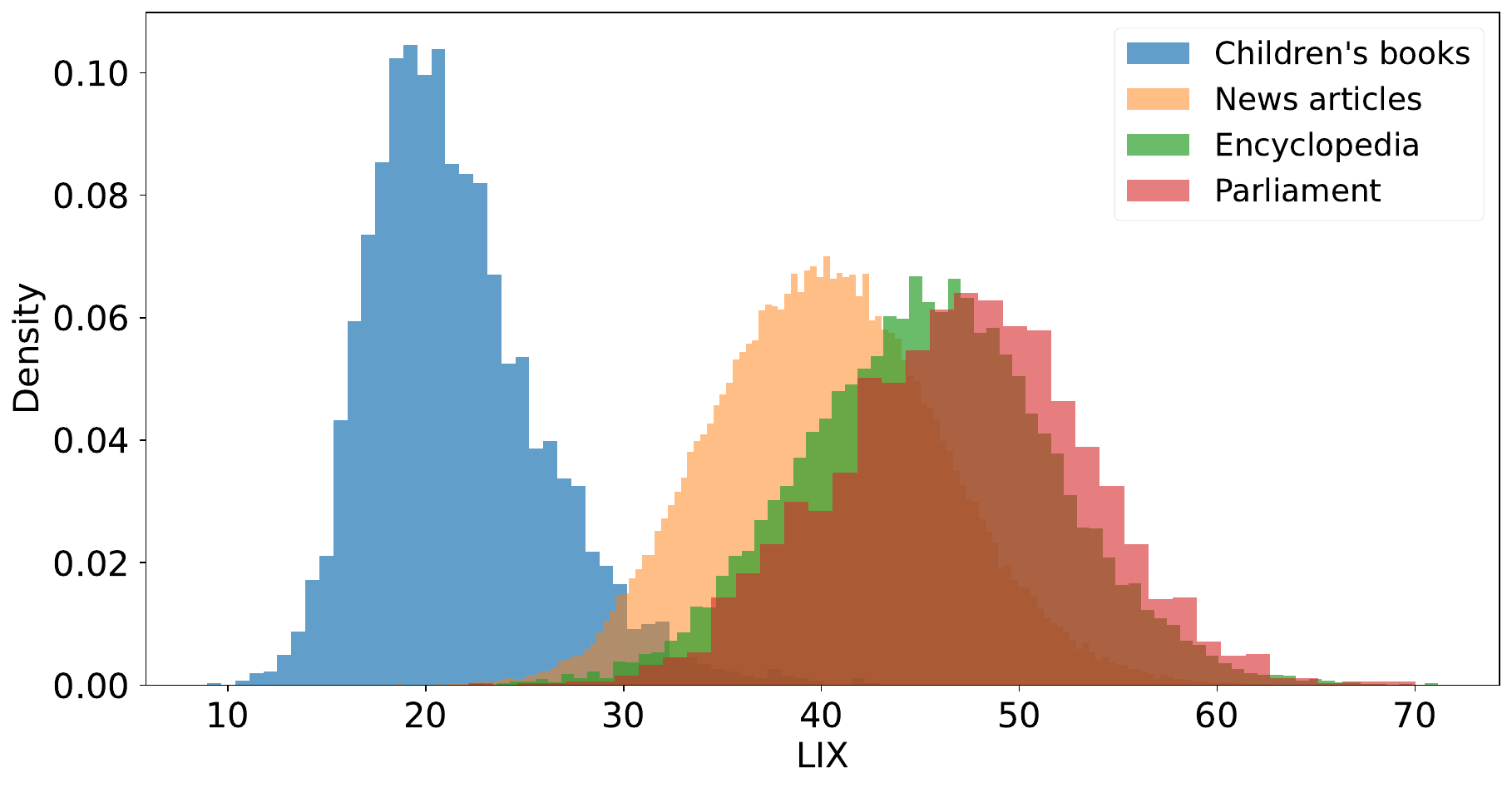}
    \caption{The distribution of LIX scores of texts from four different corpora. From left to right: children's books, news articles, encyclopedia entries, and legislative texts from the Norwegian parliament.}
    \label{fig:corpora}
\end{figure*}
\begin{table}[]
\normalsize
  \centering
    \begin{tabular}{@{}lcc@{}}
    \toprule
    \textbf{Pair} & \textbf{Statistic} \\
    \midrule
    Children - News & $0.92$\\ 
    Children - Encyclopedia & $0.95$\\ 
    Children - Parliament & $0.97$\\ 
    News - Encyclopedia & $0.33$\\ 
    News - Parliament & $0.43$\\ 
    Encyclopedia - Parliament & $0.11$\\ 
    \hline
    \end{tabular}
    \caption{Results from a two-sample Kolmogorov-Smirnov test between corpora pairs.}
    \label{tab:significance_test}
\end{table}

To verify that the LIX index is indeed able to separate our four collected corpora into our four discrete classes of complexity, we conduct a ~ two-sample Kolmogorov-Smirnov test to verify that the samples are not from the same distribution, e.g. that they are different from each other with respect to the complexity metric. We calculate the statistics between all pairs using the two-tailed version of the test. The results from this test can be seen in \cref{tab:significance_test}. 

For this statistic, all tests are significant with a significance level of $0.01$. This confirms the overall picture painted by the raw distributions of the LIX scores from the four corpora, which can be seen in \cref{fig:corpora}. The children's books corpus receives considerably lower LIX scores than the three other categories. The distances between the texts from this category and the others are high ($.92, .95, .97$), while the distance between texts from the encyclopedia and parliament corpora is low ($.11$) but still significant. We also note that the distribution of texts in the children's corpus is more narrow compared to the three others. A possible interpretation of this observation is that texts within this category are more homogeneous. 

\subsection{Estimating lexical complexity}
\begin{figure*}
    \centering
    \includegraphics[width=0.7\textwidth]{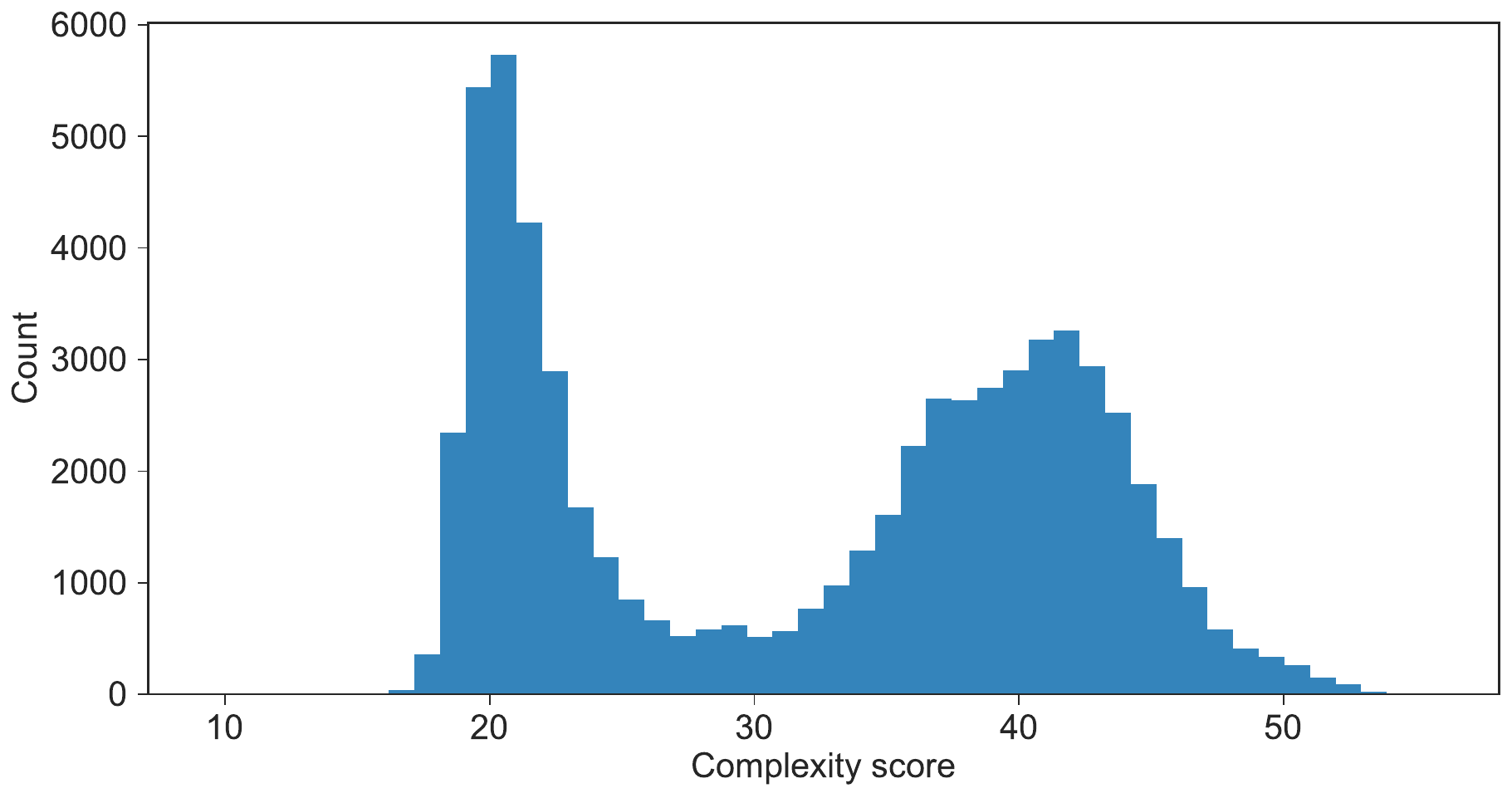}
    \caption{Normalised LIX scores for all content words in our corpus ($n=64\,071$)}
    \label{fig:normalised}
\end{figure*}
As the results from \cref{sec:doc_level} show that the LIX ~ index can separate the different levels of complexity, we can now calculate a measure for lexical complexity given \cref{assump:1} and \cref{assump:2}.

We create an inverted index where the search key is a lemma and the value is a list of all the documents in which this lemma occurs. The frequency of a lemma within a single document is disregarded and is counted as present/not present so that terms with high domain-specificity do not conflate the aggregates. The inverted index is created over documents from all four corpora. As the LIX index does not distribute values evenly across a closed interval, we apply feature scaling on the medians using a custom normalization function, arriving at a final lexical complexity score for a given lemma:

\begin{equation*}
    CS(lemma) = x * (1 - (\frac{n}{m})),
\end{equation*}

where $x$ is the median LIX score of the $n$ documents in which this lemma occurs, and $m$ is the total number of documents. This is essentially discounting the median with the proportion of the documents in which this lemma occurs. The distribution of complexity scores for a sample of lemmas can be seen in \cref{fig:normalised}. The intuition behind doing this is that we want to push high-frequency words to the left of the distribution. We elaborate on the relationship between complexity and frequency in \cref{sec:lin_freq}. We note that we only calculate complexity scores for content words with the following parts of speech: nouns, verbs, adjectives, and adverbs. 

\section{Generality}
As the final complexity scores are not based on data from one specific domain, but rather on four rather different categories, our two-step approach is not limited to texts of a specific type. Furthermore, as it does not require any pre-annotated corpora, it can be reproduced for any language given that one has access to a reliable document-level complexity measurement. We use the LIX score in this work, as it was originally developed for Swedish --- a language that is relatively similar to Norwegian with respect to vocabulary construction, compounds, morphology, and syntax --- but any language-specific complexity measure would suffice under the same conditions. 

To test for the generality of the approach, we also demonstrate the same procedure using the Coleman-Liau index, developed for English. Norwegian and English are similar enough for this to be interesting, both are Germanic languages and have long-standing cultural connections, but still different with respect to some key features that influence the average length of words. For example, English uses determiners to mark definiteness, but in Norwegian this is marked morphologically, by suffixation that increases the average word length. As we do not have access to the raw texts in the children corpus, this test is limited to the news, encyclopedia, and parliament corpora.

\begin{table}[]
\normalsize
  \centering
    \begin{tabular}{@{}lcc@{}}
    \toprule
    \textbf{Pair} & \textbf{Statistic}  & \textbf{Sig.} \\
    \midrule
    News - Encyclopedia & $0.40$ & $0.0$ \\ 
    News - Parliament & $0.39$ & $0.0$\\ 
    Encyclopedia - Parliament & $0.03$ & $0.005$ \\ 
    \hline
    \end{tabular}
    \caption{Results from a two-sample Kolmogorov-Smirnov test between corpora pairs for the Coleman-Liau test.}
    \label{tab:coleman_test}
\end{table}

As with the LIX score, the Coleman-Liau index quantifies the readability of a text based on directly observable features \citep{coleman1975computer}. The $CLI$ score is defined as:
\begin{equation*}
    0.0588 \times L - 0.296 * S - 15.8,
\end{equation*}
where $L$ is the average number of letters per 100 words and S is the average number of sentences per 100 words. 

We test this measurement on our corpora using the same pre-processing and normalization pipeline as described in section \cref{sec:pre_proc}. As with the LIX method, the Coleman-Liau is able to distinguish between the different categories of texts, but with lower margins. All pair-wise distances are still significant using the two-sample Kolmogorov-Smirnov test, but between the encyclopedia and news domain, there is considerably less distance than with the LIX ($0.03$ v.s $0.11$). The result of these tests can be seen in \cref{tab:coleman_test}.

This test concludes that the choice of complexity measure needs to account for the linguistic features of the language at hand. If not, it is not possible to use \cref{assump:2} to produce lexical scores.

\section{Qualitative evaluation}
\begin{table*}[ht]
\normalsize
\centering
\begin{tabular}{@{}lccc@{}}
    \toprule
    \textbf{Lemma} & \textbf{}  & \textbf{CS} & \# \\
    \midrule
    \norex{bety} & \eng{means} &  $33.09$  & $18330$\\ 
    \norex{resultere} & \eng{result} &  $40.77$ & $1675$ \\ 
    \norex{\textbf{medføre}} & \eng{cause} &  $41.67$ & $3984$\\
    \norex{tilsi} & \eng{entail} &  $41.88$ & $2190$ \\ 
    \norex{vanskeliggjøre} & \eng{convolute} & $46.07$ & $199$ \\
    \norex{nødvendiggjøre} & \eng{necessitate} & $47.17$ & $22$ \\
    \hline
    \end{tabular}
\quad
\begin{tabular}{@{}lccc@{}}
    \toprule
    \textbf{Lemma} & \textbf{}  & \textbf{CS} & \# \\
    \midrule
    \norex{tretthet} & \eng{tiredness} & $21.55$ & $228$ \\
    \norex{skyldfølelse} & \eng{guilt} &  $29.64$ & $198$ \\ 
    \norex{smerte} & \eng{pain} &  $31.71$ & $2685$ \\ 
    \norex{irritasjon} & \eng{irritation} & $32.98$ & $435$ \\
    \norex{stress} & \eng{stress} &  $34.86$ & $784$\\
    \norex{\textbf{ubehag}} & \eng{discomfort} & $38.37$ & $392$ \\
    \hline
    \end{tabular}
    \caption{Substitution suggestions of the lemmatised content words from \cref{exmp:medføre} (left, verb) and \cref{exmp:ubehag} (right, noun) ordered by their Norwegian complexity score. The boldfaced word is the word originally used in the assessment tool. \# denotes the document frequency.}
    \label{tab:eval_schema}
\end{table*}

\begin{table*}[ht]
\normalsize
\centering
\begin{tabular}{@{}lccc@{}}
    \toprule
    \textbf{Lemma} & \textbf{}  & \textbf{CS} & \# \\
    \midrule
    \norex{\textbf{betrakte}} & \eng{consider} & $35.77$ & $1607$ \\
    \norex{oppfatte} & \eng{perceive} &  $38.57$  & $3733$\\ 
    \norex{anse} & \eng{regard} &  $41.60$ & $4657$ \\ 
    \norex{definere} & \eng{define} &  $41.76$ & $2002$ \\ 
    \norex{betegne} & \eng{designate} &  $42.14$ & $1815$\\
    \norex{karakterisere} & \eng{characterise} & $42.73$ & $1145$ \\
    \hline
    \end{tabular}
\quad
\begin{tabular}{@{}lccc@{}}
    \toprule
    \textbf{Lemma} & \textbf{}  & \textbf{CS} & \# \\
    \midrule
    \norex{viss} & \eng{certain} &  $34.72$ & $7009$ \\ 
    \norex{\textbf{bestemt}} & \eng{specific} & $37.40$ & $3851$ \\
    \norex{enkelt} & \eng{some} &  $38.51$ & $10316$ \\ 
    \norex{akeseptabel} & \eng{acceptable} &  $41.16$ & $732$\\
    \norex{nøytrale} & \eng{neutral} & $41.61$ & $827$ \\
    \norex{spesifikk} & \eng{specific} & $41.98$ & $1142$ \\
    \hline
    \end{tabular}
    \caption{Substitution suggestions of the lemmatized content words from \cref{exmp_betrakte} (left, verb) and \cref{exmp:bestemte} (right, adjective) ordered by their Norwegian complexity score.}
    \label{tab:eval_schema_EDE}
\end{table*}

As we do not have access to any reference measurements at the lemma level, we evaluate our complexity score by assessing its usage within our target domain: mental health assessments. We examine assessment inventories developed for adolescents and adults and identify words that could be simplified. We limit our investigation to content words: verbs, nouns, adverbs, and adjectives. For each identified content word, we sample related words that could have been used in the same context and compare their complexity score. These suggested substitutes are sampled using a word-embedding model \citep{mikolov2013distributed}. This model was trained predominately on Norwegian web archives, with an embedding dimension of $100$ and a context window of $5$. For a candidate lemma, we retrieve the top similar words from the model and remove words with a different word sense. Similar words are retrieved by extracting the top N words with the lowest cosine distance from the target word. This will also make it easier for non-Norwegian readers to compare the complexity scores relative to the semantic similarity.

We note that this evaluation contains examples from a small inventory and is primarily meant as a proof-of-concept of how our complexity metric could be used in practice. An extensive qualitative analysis would require real patient feedback, which is beyond the scope of this work.

\subsection{Y-BOCS}
We evaluate our approach on samples from the official Norwegian translation of the Yale-Brown Obsessive Compulsive Scale (Y-BOCS), published by the Norwegian Society for Cognitive Therapy. \footnote{\url{https://www.kognitiv.no/}} The target group includes both adolescents and adults. For each example, we look at one of the most defining content words. The candidate is marked with a bold typeface in the Norwegian version.   

\begin{exmp} \label{exmp:medføre}
    \norex{...\textbf{medfører} betydelig svekkelse I sosiale eller I arbeidsmessig utfoldelse} \eng{...causes substantial impairment in social or occupational performance}
\end{exmp}

\begin{exmp} \label{exmp:ubehag}
    \norex{Hvor mye \textbf{ubehag} medfører tvangstankene?} \eng{How much discomfort do the obsessions cause?}
\end{exmp}

\begin{figure*}%
    \centering
    \subfloat[\centering Low frequent frequency lemmas ($<5\%$)]{{\includegraphics[width=0.47\textwidth]{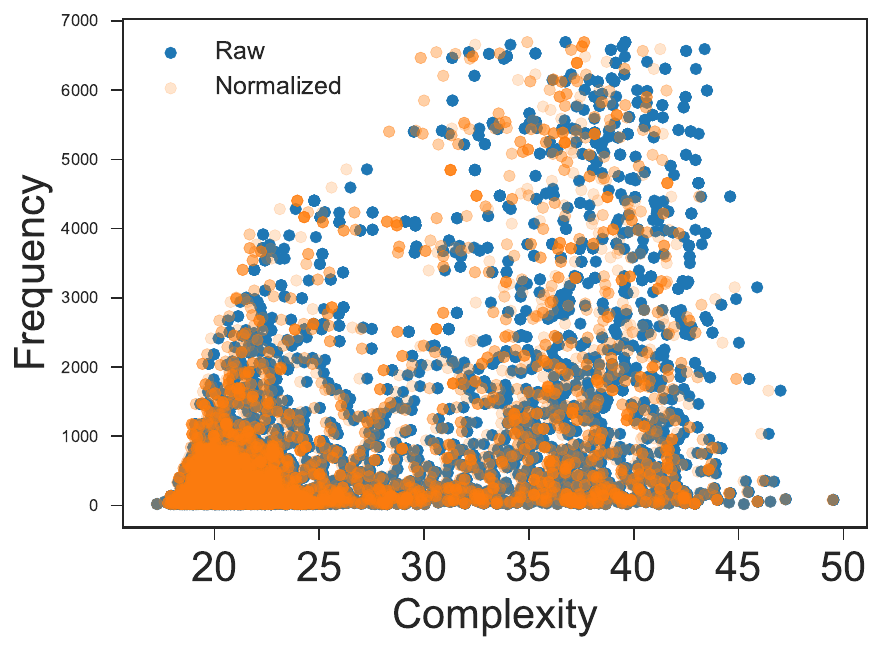}}}%
    \qquad
    \subfloat[\centering High frequent frequency lemmas ($>5\%$)]{{\includegraphics[width=0.47\textwidth]{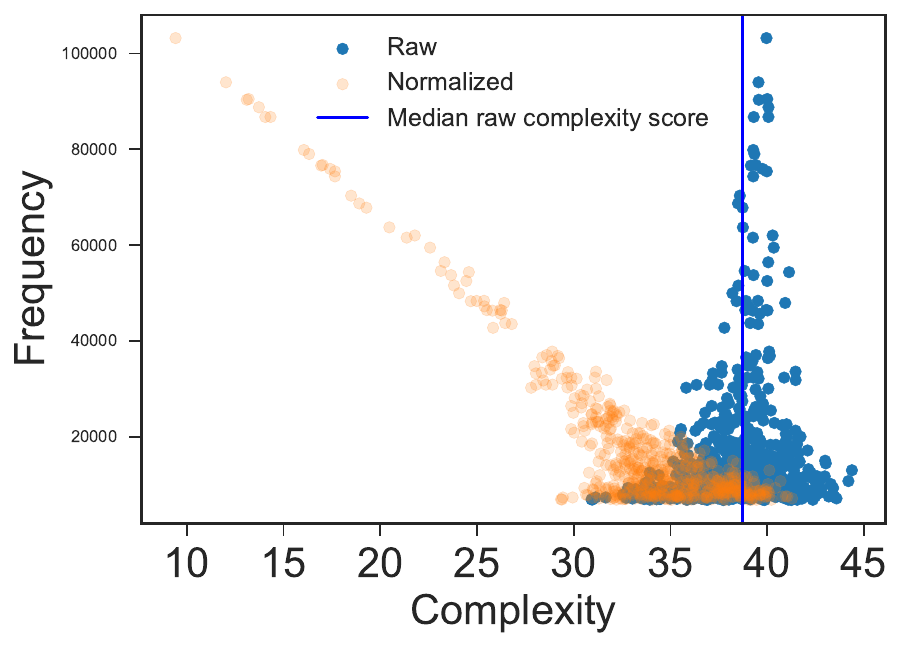}}}%
    \caption{The relationship between frequency and complexity score after normalization for a selection of lemmas.}%
    \label{fig:lin_freq}%
\end{figure*}

\subsection{EDE-Q 6.0}

We follow the same procedure as above but on samples from the Eating Disorder Examination Questionnaire (EDE-Q 6.0). The target group is still both adolescents and adults.

\begin{exmp}\label{exmp_betrakte}
    \norex{I løpet av de siste 28 dagene, hvor mange ganger har du spist det andre ville \textbf{betraktet} som en uvanlig stor mengde mat (omstendighetene tatt i betraktning)?} \eng{In the last 28 days, how many times have you eaten what others would consider an unusually large amount of food (given the circumstances)?}
\end{exmp}

\begin{exmp}\label{exmp:bestemte}
    \norex{Har du prøvd å følge \textbf{bestemte} regler for hva eller hvordan du spiser (f.eks. en kalorigrense)...?}
    \eng{Have you tried to follow specific rules for what or how you eat (e.g. a calorie limit)...?}
\end{exmp}

\subsection{Results}

The results of applying the complexity measure to the content words and their top substitutes from the embedding model can be seen in \cref{tab:eval_schema} and \cref{tab:eval_schema_EDE}. We note that providing accurate translations of lemmas external to a context necessarily causes inaccuracies, and that the ordering of lexical complexity in Norwegian does not correspond 1-1 with the English glossary when compared side by side. 

We observe that the most frequent lemma is not necessarily the one with the lowest complexity score. Due to our normalization, we somewhat discount the effect of high-frequency terms. The relationship between complexity score and frequency is explored in more detail in \cref{sec:lin_freq}. We observe that a potential substitution of the original lemma with one of the alternatives can reduce the complexity substantially. For \cref{exmp:medføre}, substituting \norex{medføre} \eng{cause} with \norex{bety} \eng{means} reduces the lexical complexity by $\approx 20\%$. For \cref{exmp:ubehag}, a substitution to \norex{smerte} \eng{pain} reduces the lexical complexity by $\approx 17\%$. For \cref{exmp_betrakte}, a substitution of the original term with another lemma would only increase the score, while for \cref{exmp:bestemte} we can swap \norex{bestemt} \eng{specific} with \norex{viss} \eng{certain} for a $\approx 7\%$ decrease in complexity score.

It follows from our assumptions and the distributions of raw LIX scores in the four corpora that most complexity scores, even after normalization, will be centered around the mean of the document-level distributions ($\mu=40.58, \sigma=6.94$). Consequently, there will be few suggested substitutes for a given lemma that are considerably different with respect to complexity, but these could reduce the cognitive load of the assessment significantly.

\section{Analysis}

\subsection{Complexity and frequency}\label{sec:lin_freq}

In \cref{fig:lin_freq} we demonstrate how our normalization affects lemmas that appear in more than $5\%$ of the documents ($n=592$) and in less than $5\%$ ($n=63479$). Through our normalization, we are effectively enforcing a linear relationship between complexity and frequency for lemmas that occur in almost all documents. It is well known that frequency correlates with complexity, as exposure to a word influences how easily we experience it. This has been the basis for previous work, such as \citet{chen-meurers-2016-characterizing}. Since most lemmas have a complexity centered around the median, see the right side of \cref{fig:lin_freq}, we can tilt this distribution leftwards so that high-frequent lemmas are given a low complexity score. For the majority of the lemmas, we want the score to be close to their raw value (the median complexity score of the documents they appear in), as that characterizes how the lemma is typically used.  

We can confirm this effect by looking at the monotonic relationship between complexity score and frequency using Spearman's correlation coefficient, and we find that for low-frequent lemmas there is no such relationship ($n=63479, \rho=0.00, p-value=0.55$), while for the high-frequent ones, there is a strong correlation ($n=592, \rho=-0.59, p-value=0.0$). 

\subsection{Complexity, syllables and word length}\label{sec:syl_wl}
\begin{figure*}%
    \centering
    \subfloat[\centering  Complexity and word length]{{\includegraphics[width=0.47\textwidth]{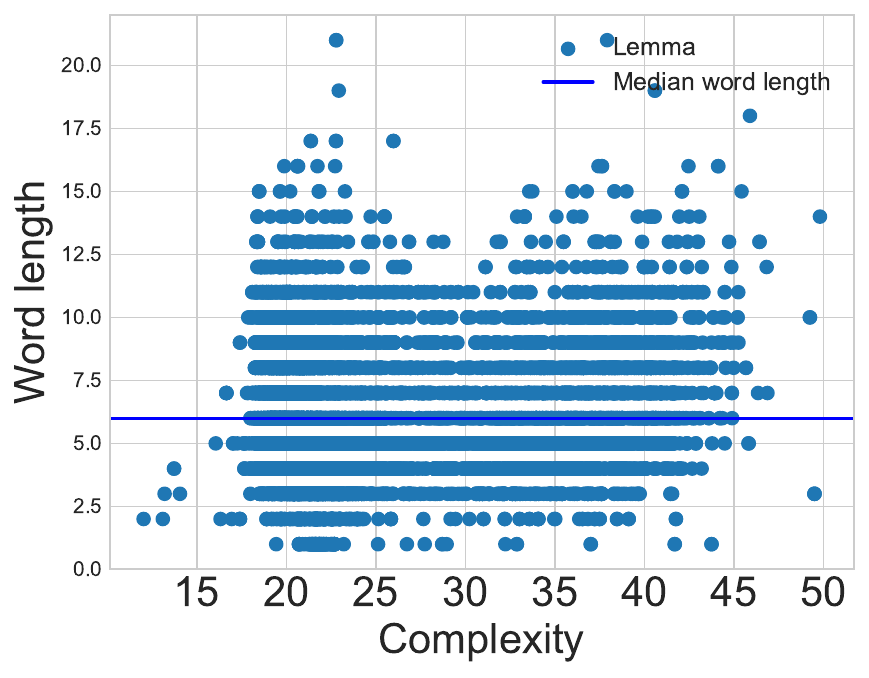}}}%
    \qquad
    \subfloat[\centering Complexity and number of syllables]{{\includegraphics[width=0.47\textwidth]{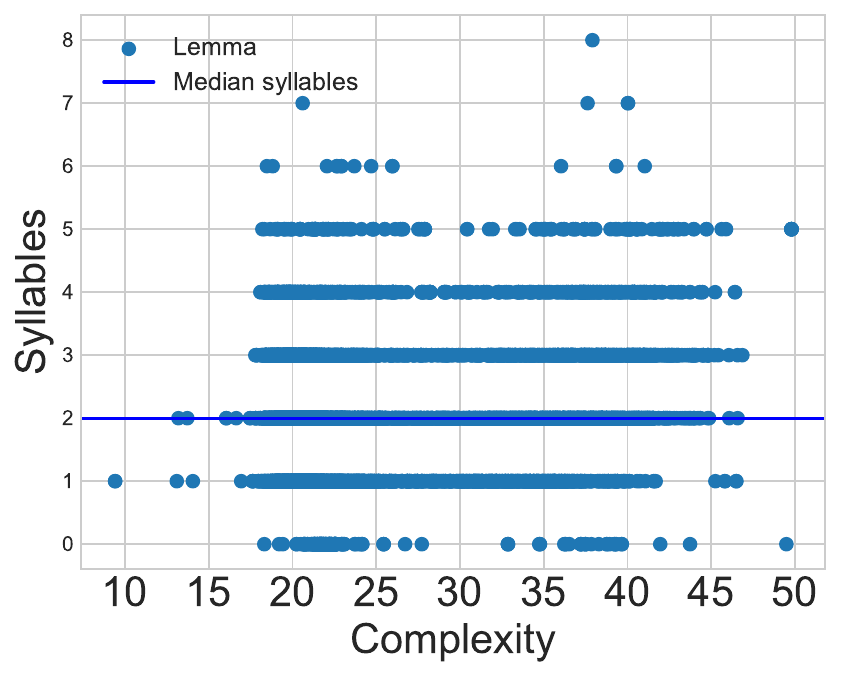}}}%
    \caption{Relationship between complexity and word-level features, from a sample of 10\,000 lemmas.}%
    \label{fig:wl_syl}%
\end{figure*}

Ignoring derivational and inflectional morphology, \cref{fig:wl_syl} shows the relationship between our complexity score, word length, and syllables. We do not observe any correlation between our scores and such word-level features. Words of different lengths are evenly spread across the complexity spectrum, and words with more syllables do not receive higher scores through our method, with the exception of short words being somewhat more frequent in the lower ranges. These words are high-frequent and are thus pushed towards the outer rims due to our normalization procedure. Syllables are counted using a custom rule-based method written for this use case, evaluated on the NLB pronunciation corpus for Bokmål and Nynorsk\footnote{\url{https://www.nb.no/sprakbanken/ressurskatalog/oai-nb-no-sbr-52/}, \url{https://www.nb.no/sprakbanken/uttaleordliste-for-nynorsk/}} at $98.2\%$ accuracy for Bokmål and $98.1\%$ for Nynorsk. We note that words of zero syllables represent non-word tokens and potential OCR errors. 

\subsection{Limitations}
Our complexity measurement has some inherent limitations. For example, if we investigate the term \norex{undulat} \eng{budgerigar}, a specific type of bird, we see that it has a lower complexity score than the general word for bird, \norex{fugl} ($19.30$ and $20.25$). The budgerigar is a common illustration in children's books but is rarely mentioned elsewhere. The word \eng{bird} is also frequent in children's books, but it can also be found in other types of texts, resulting in somewhat perplexing results. This ``budgerigar phenomena'' is a general limitation of our approach, and it follows directly from \cref{assump:1}. One could argue that \eng{bird} is an abstract concept, while \eng{budgerigar} is tangible, so our scores are justified, but if the general correlation between exposure and complexity is correct, as assumed by most, our method does not account for such cases. 

\section{Previous work}

\subsection{Lexical features in complexity metrics}

The LIX score has some history of use in Norway. \citet{liks_mercator} explores the relationship between loanwords and LIX in textbooks. \citet{golden_ordforraad} mentions LIX as a commonly used tool for indicating difficulty ( \norex{vanskelighetsgrad}), but notes that the score does not take into account the actual content side of the words \citep[p. 172]{golden_ordforraad}. Other investigations into the relationship between readability and word-level features include \citet{bunker1988}, as cited in \citet{golden_ordforraad}, who found that in a manually annotated corpus, neither word length in terms of letters nor in syllables, correlated with difficulty, but that frequency did correlate. Also, the degree of variation correlated with the observed difficulty. 
Both LIX \citep{bjornsson1968lasbarhet} and other complexity measures such as Coleman-Liau \citep{coleman1975computer} are based on easily available features, such as the number of sentences, words, and number of characters. In addition, the Flesch, Flesch-Kincaid, and Gunning Fog reading ease scores include syllables. 

While there is much work on CWI and lexical simplification, such as shared tasks \citep{yimam-etal-2018-report}, and several datasets \citeplanguageresource{shardlow-2013-cw, yimam-etal-2017-cwig3g2}, much of the focus has been on second language learners. While potentially helpful in this regard, the focus of this project is on native speakers. 

\subsection{Estimating complexity in NLP}
In NLP, the estimation of text complexity has predominately focused on the document-level. Comparable to the gist of our method, but for whole documents, is the TextEvaluator approach \citep{sheehan-etal-2013-two}. The authors first classify documents into three different genres and then generate a complexity score by applying genre-specific regression models to the texts, using US grade levels as the supervision signal. With a similar setup, \citet{flor-etal-2013-lexical} also focuses on the relationship between text complexity levels and US grade levels, but assessed by looking at \textit{lexical tightness} --- the degree to which a text tends to use words that are typically inter-associated. The authors conjecture that words that often appear in the same document, and thus are highly inter-associated, are easier to read and correspond to lower grade levels. Similar to our work, they assume that there is a relationship between word-level statistics and document complexity. More recently, a shared task at SemEval \citep{shardlow2021semeval} proposed to assign words to complexity classes from a five-step Likert scale, relying on crowd-sourced complexity labels as the supervision signal. The words were predicted in context, and it is, therefore, no surprise that the top-performing system, JustBlue \citep{bani-yaseen-etal-2021-just}, relies on a mix of pre-trained language models for the classification.


\section{Conclusion}
In this work, we motivate and develop a two-step approach for estimating lexical complexity scores from document-level distributions. Based on a version of the distributional hypothesis, we calculate the complexity score of a lemma as the median LIX index of documents in which this lemma occurs. We evaluate our approach on samples from mental health assessments and find that our approach, when coupled with semantic similarity searches, could be an important tool for health practitioners. We develop and test our approach for the Norwegian language, but we also demonstrate its generality using an English readability index. Through normalization, we can enforce a monotonic relationship for high-frequency lemmas, while maintaining a tight connection between actual usage and complexity for the majority of terms. In line with some previous work, we also find that word length and number of syllables do not correlate with our complexity score. We also identify some limitations that follow from our methodological assumptions.

\section{Limitations and future work}
Our method is based on simple, corpus-based techniques. As previously mentioned, some rare terms might receive low complexity scores as they only appear in texts that are otherwise rather easy, such as the word \norex{undulat} \eng{budgerigar} in children's books. Furthermore, complexity extends beyond the scope of a single lemma. A typical strategy for simplifying language is to replace words that are experienced as difficult with multiple words, something our method is not able to suggest. One possibility in this regard is to combine our method with generative language models. However, these models are not straightforward to implement in healthcare due to their opaqueness.  
We also acknowledge that while the LIX score is in use for Norwegian, and still shows promising results for Norwegian, it was formulated at a time when computational power was limited. The original definition of LIX involved doing a number of selections randomly and counting manually, and the components of the LIX score are selected both due to correlation with human-annotated scores, but also due to ease of computation, using features that would have low errors by counting manually. We believe further investigations into the foundations of these metrics might yield further insight into how text and lexical complexity may be calculated for Norwegian and other languages.
We urge the users of complexity measures, be it ours, the original LIX score, or others, to be vary of the fact that these only show one side of a complex problem and that care should be given when labeling text, especially individually words. We especially note that syntactic information is not explicitly contained by our metric.
The next step is to verify the applicability of our method using real patient feedback. We also hope to expand our analysis of lexical complexity from the perspective of NLP, collecting more diverse data with respect to both source and language.

\nocite{*}
\section{Bibliographical References}\label{sec:reference}

\bibliographystyle{lrec-coling2024-natbib}
\bibliography{lrec-coling2024-example}

\section{Language Resource References}
\label{lr:ref}
\bibliographystylelanguageresource{lrec-coling2024-natbib}
\bibliographylanguageresource{languageresource}

\end{document}